\documentclass{edm_article}
\usepackage{balance}

\usepackage{amsmath,amsfonts,bm, bbm}

\def\eqref#1{equation~\ref{#1}}

\def\1{\bm{1}}

\def\rvb{{\mathbf{b}}}

\def\rvy{{\mathbf{y}}}

\def\rmV{{\mathbf{V}}}
\def\rmW{{\mathbf{W}}}

\def\ve{{\bm{e}}}

\def\vh{{\bm{h}}}

\def\vx{{\bm{x}}}
\def\vy{{\bm{y}}}

\DeclareMathAlphabet{\mathsfit}{\encodingdefault}{\sfdefault}{m}{sl}
\SetMathAlphabet{\mathsfit}{bold}{\encodingdefault}{\sfdefault}{bx}{n}

\usepackage{mdframed}

\usepackage{accents}
\newlength{\dhatheight}

\usepackage{url}
\usepackage{xcolor}
\usepackage{makecell}
\usepackage{booktabs}
\usepackage{multirow}
\usepackage{subcaption}
\usepackage{amssymb}

\usepackage[flushleft]{threeparttable}
\usepackage[labelfont=bf,justification=justified,singlelinecheck=false]{caption}
\begin{document}

\title{Math Operation Embeddings for Open-ended Solution Analysis and Feedback}
\numberofauthors{1}
\author{
    Mengxue Zhang$^1$, Zichao Wang$^2$, Richard Baraniuk$^2$, Andrew Lan$^1$\thanks{This work is supported by the National Science Foundation under grant IIS-1917713.} \\
    \affaddr{$^1$University of Massachusetts Amherst, $^2$Rice University}
}

\maketitle

\begin{abstract}
Feedback on student answers and even during intermediate steps in their solutions to open-ended questions is an important element in math education. Such feedback can help students correct their errors and ultimately lead to improved learning outcomes.
Most existing approaches for automated student solution analysis and feedback require manually constructing cognitive models and anticipating student errors for each question.
This process requires significant human effort and does not scale to most questions used in homeworks and practices that do not come with this information. 
In this paper, we analyze students' step-by-step solution processes to equation solving questions in an attempt to scale up error diagnostics and feedback mechanisms developed for a small number of questions to a much larger number of questions. 
Leveraging a recent math expression encoding method, we represent each math operation applied in solution steps as a transition in the math embedding vector space. 
We use a dataset that contains student solution steps in the Cognitive Tutor system to learn implicit and explicit representations of math operations.
We explore whether these representations can i) identify math operations a student intends to perform in each solution step, regardless of whether they did it correctly or not, and ii) select the appropriate feedback type for incorrect steps. 
Experimental results show that our learned math operation representations generalize well across different data distributions.
\end{abstract}

\keywords{Embeddings, Feedback, Math expressions, Math operations}

\section{Introduction}

Math education is of crucial importance to a competitive future science, technology, engineering, and mathematics (STEM) workforce since math knowledge and skills are required in many STEM subjects \cite{limitperf}. One important way to help struggling students improve in math is to diagnose errors from student answers to math questions and deliver personalized support to help them correct these errors \cite{adams}. In short-answer questions, feedback of various types \cite{shute} can be deployed according to the specific incorrect final answers students submit, while in open-ended questions, feedback can be deployed at intermediate solution steps according to the specific actions they take and their outcomes \cite{mlp}. In traditional educational settings, this feedback process relies on teachers going over student work, identifying errors, and providing feedback \cite{kelly}, which results in a labor-intensive process and a slow feedback cycle for students. Such a setting is even more limited as a result of the COVID-19 pandemic, which introduced new barriers to face-to-face interactions between teachers and students. 

In intelligent tutoring systems, a more scalable approach to math feedback is to automatically deploy feedback based on students' final answers or certain incorrect intermediate solution steps. For example, in ASSISTments \cite{assistments}, teachers can create hints and feedback messages for specific incorrect student answers to short-answer questions that they anticipate \cite{teachersource}, which the system can automatically deploy when students submit these incorrect answers. This crowdsourcing approach efficiently scales up teachers' effort so that they can benefit a large number of students without putting in additional effort. 
In many other systems such as Cognitive Tutor \cite{ct} and Algebra Notepad \cite{orourke}, researchers use cognitive models to anticipate student errors as results of buggy production rules or insufficient knowledge on key math concepts \cite{kencor,kenmis}. They then develop corresponding feedback for intermediate solution steps in multi-step questions (e.g., those on equation solving). This cognitive model-based approach requires significant effort by domain experts and has shown to be highly effective in large-scale studies. 

However, these approaches for student feedback are still limited in their generalizability to many math questions deployed in daily homeworks and practices. For the teacher crowdsourcing approach, hint and feedback messages have to be written for each individual question (or group of questions generated from the same template with different numerical values). For the cognitive model-based approach, a rigorous solution process has to be specified for each question with annotations on the math operations that should be applied at each solution step. However, questions used in many real-world educational settings do not come with such information; teachers simply adopt them from sources such as textbooks and open education resources and assign them to students without developing corresponding feedback mechanisms. Moreover, past research has shown that a large portion of incorrect student answers cannot be anticipated by cognitive models \cite{kurt}, teachers/domain experts \cite{erickson}, or numerical simulations \cite{selent}. Therefore, it may be hard for high-quality feedback developed for questions used in intelligent tutoring systems to generalize to questions in the wild.

\subsection{Contributions}

In this paper, we develop \emph{data-driven} methods that enable us to analyze step-by-step solutions to open-ended math questions. In contrast to existing methods that rely on a \emph{top-down} approach, i.e., defining the structure of the solution process and anticipating student errors, we propose a \emph{bottom-up approach}, i.e., using learned representations of \emph{math expressions} and \emph{math operations} to predict i) math operations in student solution steps and ii) the appropriate feedback for incorrect solution steps. We restrict ourselves to the specific domain of \emph{equation solving} where the solution process consists of applying specific math operations between math expressions in consecutive steps; other sub-domains of math such as algebra word problems \cite{algebraword} and questions involving graphs and geometry \cite{scienceqa} are left as future work. Specifically, our contributions are:
\begin{itemize}
    \item First, we characterize math operations by how they \emph{transform} math expressions in the math embedding space in each solution step.
    We leverage recent work on learning \emph{math symbol embeddings} from large-scale scientific formula data \cite{forte} to encode math expressions in student solutions: each math expression is mapped to a point in the \emph{math embedding vector space}. 
    We use synthetically generated data as well as solution steps generated by real students to learn the representation of each math operation. 
    We explore several methods for learning both implicit and explicit math operation representations: a classification-based method that does not explicitly impose a structure on math operations, a linear model that assumes each operation is characterized by an \emph{additive vector} in the embedding space, and a nonlinear model where math operations live in their own, interconnected embedding spaces. 
    \item Second, we apply these math operation representation learning methods to a real-world student step-by-step solution dataset collected while student learn equation solving in an intelligent tutoring system, Cognitive Tutor \cite{ct}. 
    We validate our math operation representation learning methods via two tasks: i) predicting the specific math operation the student intended to apply in a solution step from the math expressions before and after the step and ii) predicting the appropriate feedback deployed to students from the incorrect math expressions they enter. 
    Quantitative results show that tree embedding-based math expression encoding methods outperform other encoding methods since they are able to explicitly capture the semantic and structural characteristics of math expressions. 
    They also have better generalizability across different data distributions and remain effective across different question difficulty levels and even when student solutions steps contain errors.
\end{itemize}
\subsection{Use Case}

\begin{figure}
\vspace{-.5cm}
    \centering
    \includegraphics[width=.8\linewidth]{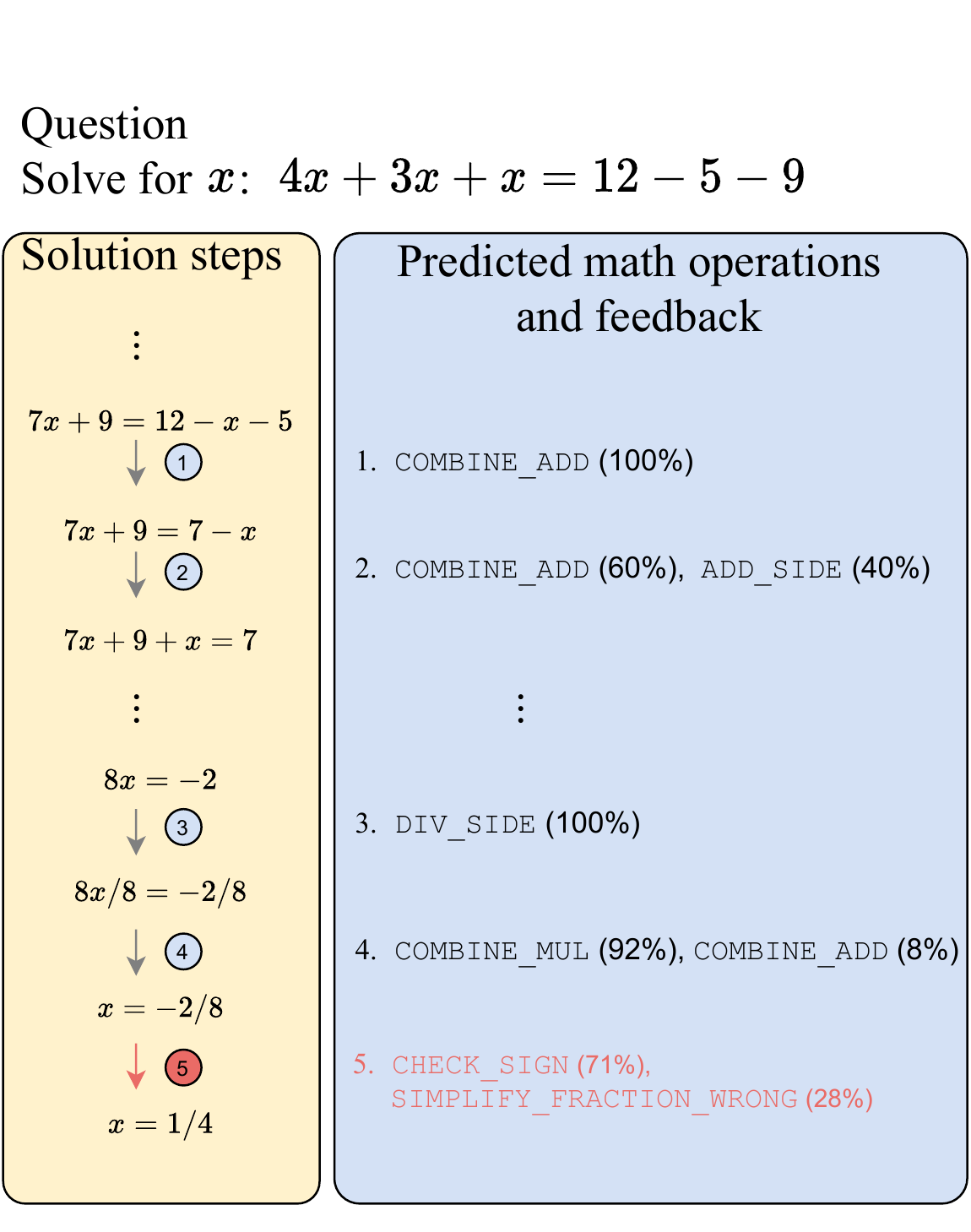}
    \caption{Demonstration of the generalizability of our math operation representations to other data sources for a solution process provided on Algebra.com. Our methods can successfully predict the math operations applied in each step and the appropriate feedback type in an incorrect step.}
    \label{fig:my_label}
\end{figure}

\sloppy
Before diving into the technical details, we first illustrate a potential use case for our math operation representation learning methods and corresponding operation/feedback classifiers. 
Our goal is to transfer expert designs in intelligent tutoring systems for math education to questions in the wild. 
Specifically, we apply the math operation representations learned from student solution steps and corresponding labels (step name, feedback message) in the highly structured Cognitive Tutor system to environments that are not highly structured. 
Figure~\ref{fig:my_label} shows the solution process to an equation solving question on Algebra.com\footnote{The original question and the solution process can be found at \url{https://www.algebra.com/algebra/homework/equations/Equations.faq.question.4872.html}.} and the corresponding math operation and feedback predictions at each step. 
We see that our math operation representation learning methods can accurately predict the math operations applied in solution steps $1$, $3$, and $4$ using the operation names provided in the Cognitive Tutor system. 
Even in step $2$ where two different math operations are combined into a single step, i.e., 
\begin{align*}
    & 7x+9=7-x \\
    & \downarrow \text{ADD }x\text{ TO BOTH SIDES} \\
    & 7x+9+x=7-x+x \\
    & \downarrow \text{COMBINE TERMS ON RIGHT SIDE}\\
    & 7x+9+x=7, 
\end{align*}

despite only training on steps in Cognitive Tutor that involve only one math operation, the classifier is able to recognize both of them with high predictive probability for both. 
We also change one of the solution steps, i.e., step $5$, to make it incorrect and test our feedback classifier. 
In this case, the classifier is able to recognize the error in this step and find the corresponding feedback types in Cognitive Tutor. 
This potential use case demonstrates the utility of our math operation representation learning methods: by transferring knowledge learned in well-designed, highly-structured systems such as Cognitive Tutor, especially on what feedback to deploy for each student error, to other domains such as online math Q\&A sites, we are \emph{scaling up} the effort domain experts put into the design of these feedback mechanisms.

\section{Related Work}
\label{sec:rw}

One related body of work in math education that studies student solution processes to identify student strategies and assess errors. Specifically, \cite{rafferty} uses inverse Bayesian planning to learn solution strategies (i.e., policies) in equation solving and capture student misunderstandings in a Markov decision process framework. 
Our work focuses on a different aspect of the solution process: the representation of the math expressions at each solution step and the modeling of the transitions between different math expressions under math operations. \cite{zoran} uses basic math operations to construct programs to understand errors that students make in their solutions to arithmetic questions. Our work focuses on equation solving, which is a more difficult problem in which students responses are are more diverse and are less structured than arithmetic calculations. \color{black}

Another related body of work focuses on learning representations of student answers to short-answer questions. \cite{kolb} analyzes incorrect student answers across multiple questions, learn representations of errors, and generalize misconception feedback across questions. Our work analyze the full math expressions in intermediate solution steps while their work represents short answers according to the frequency they occur in an answer pool. \cite{erickson} uses trained word embeddings to represent short answers for automated grading purposes. Our work focuses on learning transitions of math expressions across solution steps instead of learning representations of only the final answer. 

In domains other than math education, there exist methods for automated feedback generation, including programming \cite{piech,price,gulwani} and essays \cite{roscoe}. However, transferring these methods to math solutions is not trivial since i) open-ended math solutions are less structured than programming code and ii) data-driven representations of math symbols have not been developed until recently \cite{forte} whereas such representations have been studied for a long time in natural language processing \cite{bert,lsa,word2vec}. 

Another body of remotely-related work focuses on using computer vision techniques to identify math expressions from images for similar math expression retrieval \cite{ecmlsim}, turning hand-written math expressions into \LaTeX \,\cite{recog}, and automatically identifying and correcting student errors \cite{tencent}. These works often bypass the inherent structure of math expressions and directly use an end-to-end model for their tasks, which means that they cannot be used to analyze student knowledge. Nevertheless, these techniques can be used to build large-scale datasets containing hand-written student solutions which we can use in the future. 

\section{Background: Embedding Math \\ Expressions into Vector Spaces}
\label{sec:background}

\begin{figure*}[t!]
    \captionsetup{justification=centering}
    \centering
    \includegraphics[width=0.7\linewidth]{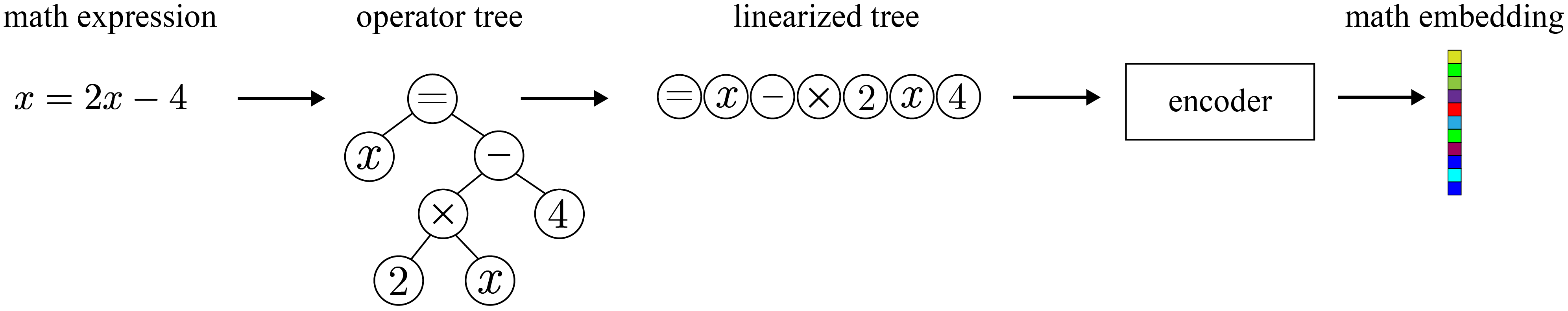}
    \centering
    \caption{Illustration of the math expression encoding method that we employ in this work.}
    \label{fig:math_encoder}
\end{figure*}

In this section, we provide an overview of a recent method that we developed to embed math expressions into a vector space, i.e., a math embedding space. 
Doing so turns discrete, symbolic math expression representations into continuous, distributed representations~\cite{bengio2003neural}, which enables us to manipulate math expressions in a manner compatible with modern machine learning methodologies. 

Our embedding method is a {\em tree-structured} encoder illustrated in Figure~\ref{fig:math_encoder}. 
The key observation is that any math expression has a corresponding symbolic {\em tree-structured} representation in the {\em operator tree} format. In the operator tree, the non-terminal (non-leaf) nodes are math operators, i.e., addition and subtraction, and terminal (leaf) nodes are numbers or variables; See Figure~\ref{fig:math_encoder} for an illustration. Thus, an operator tree explicitly captures the semantic and 
structural properties of a math expression. 

A number of existing works have demonstrated the superior performance of using operator tree representations of math expressions compared to other math expression representations in applications such as automatic math word problem solving~\cite{mwp-tree-3,mwp-tree-1,mwp-tree-2} and math formulae retrieval~\cite{tangent-s,TangentCFT,approach0-new,approach0}. 

Therefore, we built a math expression encoder that leverages the operator tree representation of math expressions. 
Specifically, during the encoding process, it first converts a math expression into its corresponding tree format, using the parser introduced in~\cite{tangent-s}. 
It then linearizes the tree by depth first search that enables us to process nodes as a sequence in which each math symbol is associated with its own trainable embedding. 
Next, it leverages positional encoding, similar to~\cite{NIPS2017_3f5ee243,tree2tree-transformer}, to retain the relative position of each node in the tree. 
The output of our encoder is a fixed-dimensional embedding vector that represents the input math expression, which we will use to learn representations of math operations for the math operation classification and feedback prediction tasks. 
We pretrain the encoder on a large corpus of math expressions extracted from Wikipedia and arXiv articles and demonstrated superior performance in reconstructing math expressions (and scientific formulae) and retrieving similar expressions. 
See the anonymized version of our work at \cite{forte}. 
We will refer to the trained encoder as the {\em math expression encoding method} in what follows.
\section{Learning Representations of \\ Math Operations}
\label{sec:method}

In this section, we detail methods we use to learn both implicit and explicit math operation representations by studying how they transform math expressions in each solution step in the math embedding space. In these methods, we leverage the math expression encoding method developed in our prior work that we reviewed above to embed math expressions into vectors and work with these embedding vectors. However, since these embeddings are trained on math expressions that are very different from those occurring in actual student solution steps, we use an additional trainable, fully-connected neural network to adapt these embeddings to our dataset, following a popular approach in natural language processing \cite{adapter}. Specifically, we have $\mathbf{e} = g_{\boldsymbol{\gamma}}(\mathbf{m})$ where $\mathbf{m}$ and $\mathbf{e}$ are the embedded vector of a math expression in our dataset before and after the adaptation, respectively. $\boldsymbol{\gamma}$ denotes the set of parameters in the fully-connected network that we will learn during the training process. 

We define a step in a student's solution to open-ended math questions as a tuple $(\mathcal{E}_1,\mathcal{E}_2,z)$, where $z \in \mathbf{Z}$ is the math operation applied in this step, with $\mathcal{Z}$ denoting the set of possible math operations. $\mathcal{E}_1 \in \mathbf{E}$ and $\mathcal{E}_2 \in \mathbf{E}$ denote the math expressions involved in this step before and after applying this math operation, i.e., the step can be expressed as 
$\mathcal{E}_1 \stackrel{z}{\longrightarrow} \mathcal{E}_2$. $\mathbf{E}$ denotes the set of all unique math expressions (across all steps in a dataset). For simplicity, we assume that only one math operation is applied in each step; an extension to cases where multiple math operations is trivial and will be discussed in what follows. $\mathbf{e}_1 \in \mathbb{R}^D$ and $\mathbf{e}_2 \in \mathbb{R}^D$ are the fine-tuned embedding vectors that correspond to math expressions $\mathcal{E}_1$ and $\mathcal{E}_2$, respectively, where $D$ is the dimension of the embedding.

\subsection{Math Operation Classification}
\label{sec:classification}

The first task we will study in this paper is to classify the math operation applied in a solution step given the math expression embeddings before and after appliying it, $\mathbf{e}_1$ and $\mathbf{e}_2$. The same notations and approaches also apply to our second task, feedback classification. This task can simply be solved using a supervised learning method, e.g., a regression model where the predicted probability of predicting a math operation $\hat{z}$ is given by
\begin{align*}
    p(\hat{z} = z) = \text{softmax}(\mathbf{v}_z^T [\mathbf{e}_1^T, \; \mathbf{e}_2^T]^T),
\end{align*}
where $\text{softmax}(\cdot)$ is the softmax function for multi-label classification \cite{dlbook}. $\mathbf{v}_z$ is a parameter vector associated with each math operation $z$, which is used to compute an inner product with the concatenation of $\mathbf{e}_1$ and $\mathbf{e}_2$ before being fed into the softmax function.
On a training dataset with given tuples $(\mathbf{e}_1, \mathbf{e}_2,z)$, we can learn the parameters ($\mathbf{v}_z$) by minimizing the cross-entropy loss \cite{dlbook} between the predicted math operation $\hat{z}$ and the actual math operation. 
This approach can be seen as learning \emph{implicit} representations of math expressions since they are captured by the classifier parameters. 

\begin{figure}
    \centering
    \includegraphics[width=.9\columnwidth]{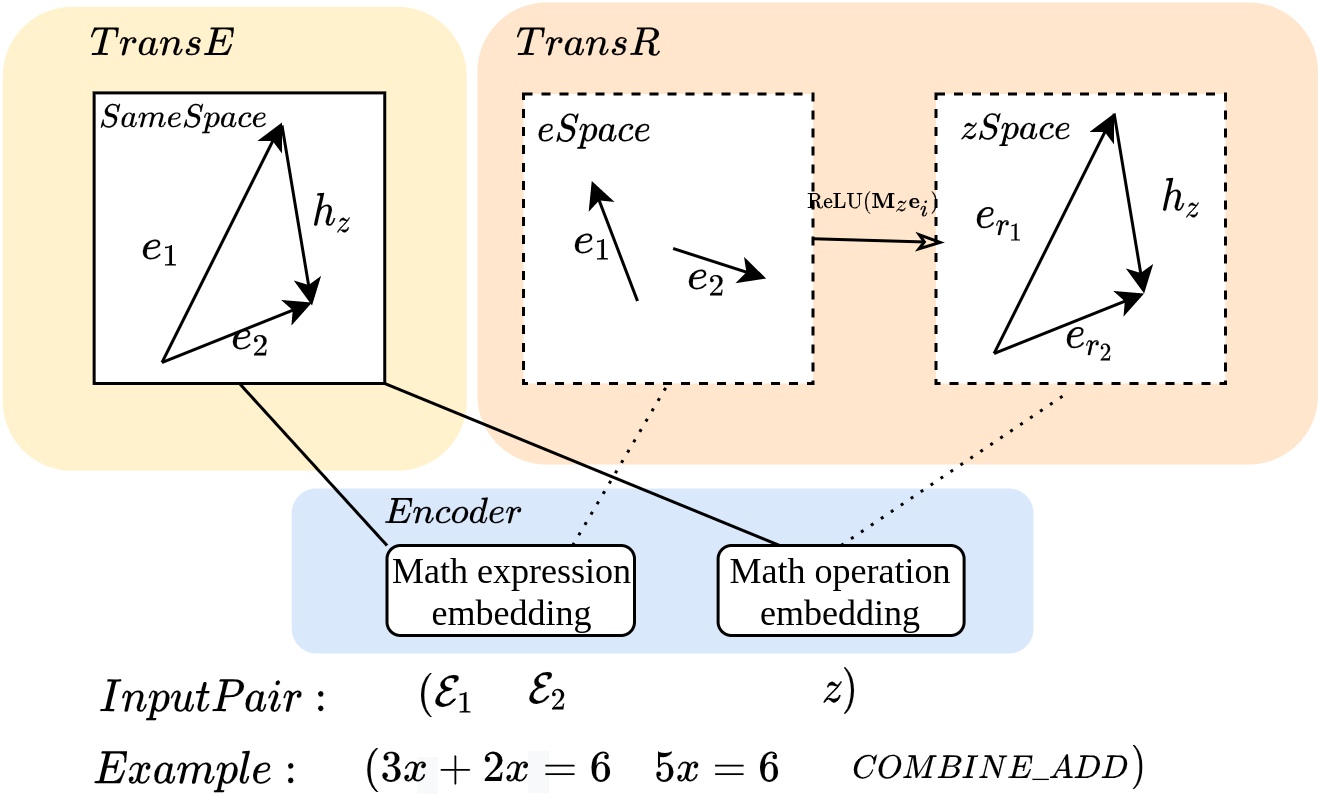}
    \caption{Illustration of the TransE and TransR frameworks. TransE puts the embeddings of equations $e_1$, $e_2$, and math operation $z$ in the same embedding space, whereas TransR puts them in their own embedding spaces.}
    \label{fig:transE}
\end{figure}

\subsection{Learning Math Operation \\ Representations}
\label{learning_math_op_rep}
The classification approach we detailed above can help us classify the math operation applied in a solution step but falls short on learning \emph{explicit} representations of math operations. The latter is important, however, to help us understand students' math solution processes and diagnose their errors. We now detail a series of methods for us to learn explicit representations of math operations. 

\subsubsection{Translating embeddings}
\sloppy
We will leverage the translating embedding (TransE) framework \cite{transE} that has found success in embedding entities and characterizing relationships between entities in multi-relational data. 
Our key assumption here in this framework is that math operations are \emph{linear and additive}, i.e., the relationship between math expressions before and after a math expression satisfy
\begin{align*}
    \mathbf{e}_2 \approx \mathbf{e}_1 + \mathbf{h}_z, 
\end{align*}
where $\mathbf{h}_z \in \mathbb{R}^D$ is the embedding of the \emph{math operation} $z$. In other words, we assume that the effect of a math operation is characterized by the difference in the embedding vectors between the math expressions before and after it in a single step; adding it to the embedded vector of $\mathcal{E}_1$ results in the embedded vector of $\mathcal{E}_2$ after the step. 

To learn these math operation embeddings from data, we use two loss functions. The first loss function promotes this linear and additive relationship between embeddings of the math expressions and operations on the training data. To this end, we define a distance function as $d(\mathbf{e}_1,\mathbf{e}_2,\mathbf{h}_z) = \|\mathbf{e}_1 + \mathbf{h}_z - \mathbf{e}_2\|_2^2$ and define the loss function as
\begin{align*}
    L_1 = \sum_{(\mathcal{E}_1,\mathcal{E}_2,z)} d(\mathbf{e}_1,\mathbf{e}_2,\mathbf{h}_z).
\end{align*}

The second loss function pushes counterfeit step tuples that are generated by replacing elements in an observed step tuple with other ones in the dataset to not satisfy the aforementioned linear and additive relationship. To this end, we minimize the pairwise marginal distance ranking-based loss given by
\begin{align*}
    L_2 & = \sum_{(\mathcal{E}_1,\mathcal{E}_2,z) \in \mathbf{S}} \, \sum_{(\mathcal{E}_1',\mathcal{E}_2',z') \in \mathbf{S}'_{(\mathcal{E}_1,\mathcal{E}_2,z)}} \\
    & \quad [\gamma + d(\mathbf{e}_1,\mathbf{e}_2,\mathbf{h}_z) - d(\mathbf{e}_1',\mathbf{e}_2',\mathbf{h}_z')]_+,
\end{align*}
where $[x]_+ = x$ when $x > 0$ and $0$ otherwise and $\gamma > 0$ is a hyper-parameter that controls the margin of the distance ranking. $\mathbf{S}$ denotes the set of steps in the dataset and 
$\mathbf{S}'_{(\mathcal{E}_1,\mathcal{E}_2,z)}$ is a set of counterfeit steps that are perturbed versions of the actual step $(\mathcal{E}_1,\mathcal{E}_2,z)$, generated by randomly replacing one of the triplet elements in the step by a different math expression or math operation from another step, i.e., 
\begin{align*}
     & \mathbf{S}'_{(\mathcal{E}_1,\mathcal{E}_2,z)} \sim A \cup B \cup C, \\
     \text{where    } & A = \{ (\mathcal{E}_1',\mathcal{E}_2,z): \mathcal{E}_1' \neq \mathcal{E}_1 \in \mathbf{E} \} \\
     & B = \{ (\mathcal{E}_1,\mathcal{E}_2',z): \mathcal{E}_2' \neq \mathcal{E}_2 \in \mathbf{E} \}\\
     & C = \{ (\mathcal{E}_1,\mathcal{E}_2,z'): z' \neq z \in \mathbf{Z} \}.
\end{align*}
Intuitively speaking, our objective encourages the distance function calculated on an actual tuple in the dataset to be smaller than that calculated on a perturbed version of it. Figure~\ref{fig:transE} illustrates the whole process.

The final loss function that we minimize is simply the combination of these two loss functions as $L = L_1 + L_2$. Using the learned embeddings of each math operation, we can classify them from the math expressions $\mathcal{E}_1$ and $\mathcal{E}_2$ using the nearest neighbor classifier, i.e., $\hat{z} = \text{argmin}_z d(\mathbf{e}_1,\mathbf{e}_2,\mathbf{h}_z)$. 

\subsubsection{Learning Entity and Relation Embeddings}
\sloppy
Despite potentially exhibiting excellent interpretability, TransE's assumption that math operations are linear and additive in the math expression embedding space may be too restrictive. 
This assumption puts math operations are vectors in the same latent space where similar math expressions will be close to each other.
However, different math operations are fundamentally different and can transform the same math expression into dramatically different math expressions that are far apart in the embedding space.
For example, different math operations can focus on transforming different parts of the same math expression. 
The steps $(3 + 5 + 2x = x + 1,\,\, 8 + 2x = x + 1,\, \text{combine similar terms})$ and $(3 + 5 + 2x = x + 1,\,\, 3 + 5 + 2x - x = x + 1 - x,\, \text{subtract from each side})$ have the same starting math expression $\mathcal{E}_1$. 
In the first step, only similar terms on the left hand side of the equation are combined, regardless of the other side of the equation. 
In the second step, we subtracted $x$ from both sides of the equation, which is a consequence of the equality symbol in the equation, which means that subtracting the same term on both sides of the equation but not what exactly is on each side. 
Therefore, TransE's linear and additive assumption means that the resulting $\mathcal{E}_2$ in these steps will be very different due to the different math operations applied, which conflicts with the observation that they are very similar. 
To address this limitation, we explore the Learning Entity and Relation Embeddings (TransR) \cite{lin2015learning} model, 
which models math expressions and math operations in different spaces, i.e., there will be a shared embedding space for all math expressions but separate relation spaces for different math operations. 

TransR learns the embeddings of math operations by projecting them to their corresponding relation spaces and then learning translations between those projected expressions. 
For each math operation $z$, we set a projection matrix $\mathbf{M}_z \in \mathbb{R}^{D \times D}$ that projects a math expression to its relation space. 
To make this projection nonlinear, we apply the rectified linear unit (ReLU) activation function \cite{dlbook} to it and define the corresponding distance function as
\begin{align*}
    d_z(\mathbf{e}_1, \mathbf{e}_2, \mathbf{h}_z) = \|\text{ReLU}(\mathbf{M}_z\mathbf{e}_1) + \mathbf{h}_z - \text{ReLU}(\mathbf{M}_z\mathbf{e}_2)\|_2^2.
\end{align*}
Correspondingly, the two loss functions in the TransR framework are given by
\begin{align*}
    L_1 & = \sum_{(\mathcal{E}_1,\mathcal{E}_2,z)} d_z(\mathbf{e}_1,\mathbf{e}_2,\mathbf{h}_z), \\
    L_2 & = \sum_{(\mathcal{E}_1,\mathcal{E}_2,z) \in \mathbf{S}} \, \sum_{(\mathcal{E}_1',\mathcal{E}_2',z') \in \mathbf{S}'_{(\mathcal{E}_1,\mathcal{E}_2,z)}} \\
    & \quad [\gamma + d_z(\mathbf{e}_1,\mathbf{e}_2,\mathbf{h}_z) - d_z(\mathbf{e}_1',\mathbf{e}_2',\mathbf{h}_z')]_+.
\end{align*}
The projection matrices $\mathbf{M}_z, \, \forall z \in \mathbf{Z}$ are included as part of the trainable parameters. The rest of the training and resulting math operation classification procedure remains unchanged from the TransE framework.
\begin{table*}[t]
    \centering
    \resizebox{1.7\columnwidth}{!}{
        \begin{tabular}{l|l|l}
            \toprule
            \textbf{Step (Math operation)} & \textbf{Description} & \textbf{Example} \\
            \midrule
             COMBINE\_ADD & combine two similar terms with add/sub operator & $3x+2x \rightarrow 5x$ \\ \hline
             COMBINE\_MUL & combine two similar terms with multiply/divide operator & $x*x \rightarrow x^2$ \\\hline 
             ADD\_SIDE & add a math term on each side & $x=1 \rightarrow x+1 = 1+1$ \\ \hline
             SUB\_SIDE & subtract a math term on each side & $x=1 \rightarrow x-1 = 1-1$ \\ \hline
             MUL\_SIDE & multiply a math term on each side & $x=1 \rightarrow x*2 = 2$ \\ \hline
             DIV\_SIDE & divide a math term on each side & $x=1 \rightarrow x/2 = 1/2$ \\ \hline
             DISTRIBUTE & distribute(expand) the terms & $(x+1)x \rightarrow x*x + x$ \\
             \bottomrule
        \end{tabular}
    }
    \caption{\centering Detailed descriptions and examples for each math operation in the CogTutor dataset.}
    \label{table:steps}
\end{table*}

\section{Experiments}
\label{sec:exp}

We now detail a series of quantitative and qualitative experiments that we have conducted to validate the learned representations of math operations. Using the Cognitive Tutor 2010 equation solving (CogTutor) dataset,\footnote{\url{https://pslcdatashop.web.cmu.edu/DatasetInfo?datasetId=660}} we focus on two tasks: i) classifying the math operation a student applies in a solution step and ii) classifying the feedback category corresponding to certain types of incorrect steps, from the math expressions the student enters before and after the step.

\subsection{Dataset}
We use the CogTutor dataset which we accessed via the PSLC DataShop \cite{pslc}. The dataset contains detailed tutor logs generated as students in a school use the Cognitive Tutor system \cite{ct} for their Algebra I class. These logs contain the students' step-by-step solutions to equation solving problems, where each step is a tuple with three elements: a \emph{math expression} $\mathcal{E}_1$ at the beginning of the step, the \emph{step name} $z$, i.e., the math operation the student selected to apply to this math expression, and the resulting math expression $\mathcal{E}_2$ after the step. 
Students can select math operations from a built-in list in Cognitive Tutor: COMBIN\_ADD, COMBINE\_MUL, ADD\_SIDE, SUB\_SIDE, MUL\_SIDE, DIV\_SIDE, and DISTRIBUTE; see Table~\ref{table:steps} for an illustration of these operations and some examples of the corresponding math operations before and after them in a step. 

There are a total of $50,406$ steps in this dataset that can be further divided into three subsets according to their outcomes: \texttt{OK} ($43,413$ steps), \texttt{ERROR} ($6,377$ steps), and \texttt{BUG} ($5,744$ steps).
The \texttt{OK} subset contains steps that are correct, i.e., the student both selected the correct math operation and arrived at the correct math expression. 
The \texttt{BUG} and \texttt{ERROR} subsets contain incorrect student steps, either because the operation they selected was incorrect or because they selected the correct operation but did not apply it correctly, i.e., arriving at an incorrect math operation after the step. 
The difference between these two subsets is that \texttt{BUG} contains steps that fit one of the predefined error templates in the Cognitive Tutor system; in this case, the system can automatically diagnose the error and deploy a predefined feedback. 
On the other hand, \texttt{ERROR} contains incorrect steps that Cognitive Tutor could not automatically diagnose the underlying error. 
The \texttt{OK} subset can be further split into six predefined difficulty levels (named as ES\_01,ES\_02, ES\_03 ,ES\_04, ES\_05, and ES\_07), with $2,068$, $7,546$, $8,183$, $13,393$, $5,484$, and $2,801$ steps, respectively. We do not further split the \texttt{BUG} and \texttt{ERROR} subsets for the math operation classification task due to their limited sizes. 

To learn the representation of math operations, we need examples of how they transform one math expression into another. 
However, the CogTutor dataset may not contain enough data that is rich in both quantity and diversity for neural network-based models to learn from. 
Therefore, we designed a synthetic data generator stemming from the math question answering dataset created by DeepMind \cite{saxton2019analysing}. 
The generator can generate steps by first generating the initial math expression and then applying math operations listed in Table~\ref{table:steps} to arrive at a resulting math expression. 
We have full control over the generated steps through the entropy, degree, and flip parameters. 
Increasing entropy introduces more complexity to the math expressions as numerical constants generated get larger. 
Increasing the degree parameter introduces monomials of higher degrees and also adds more terms in the math expression.
Finally, the flip parameter allows us to control which side of an equation has a higher chance to be more complicated than the other.
Tuning these parameters within this flexible synthetic data generation method enables us to generate a large amount of steps that closely resembles those in the CogTutor dataset. 

\subsection{Methods}

To fully evaluate the effectiveness of our math operation representations, we also experiment with two other ways of encoding math expressions commonly used in natural language processing tasks, in addition to the tree embedding-based and translation-based encoder that we introduced in Section~\ref{learning_math_op_rep}. 
These two encoders include a gated recurrent unit (GRU)-based encoder~\cite{gru-cls} and a convolutional neural network (CNN)-based encoder~\cite{cnn-cls}; we will use the output of these encoders to replace $[\mathbf{e}_1^T, \; \mathbf{e}_2^T]^T$ as input to the classifier detailed in Section~\ref{sec:classification}. 

Specifically, these two encoders first concatenates the two math expressions before and after the step, i.e., $\mathcal{E} = [\mathcal{E}_1, \mathcal{E}_2]$. 
For each character $x_t$ in $\mathcal{E}$, we compute its embedding 
\begin{align*}
    \vx_t &= \rmW^T {\rm onehot}(x_t)\,,
\end{align*}
where $\rmW$ is a trainable embedding matrix.
Using these character embeddings, the {GRU encoder} computes 
\begin{align*}
    \vh_t &= {\rm GRU}_\theta (\vx_t, \vh_{t-1})\,,
\end{align*}
where $\theta$ represents all the trainable parameters in GRU. We then replace $[\mathbf{e}_1^T, \; \mathbf{e}_2^T]^T$ with $\vh_T$ as input to the classifier where $T$ is the total number of characters in $\mathcal{E}$.
Similarly, the {CNN encoder} computes 
\begin{align*}
    \vh &= {\rm max\_pool}({\rm CNN}_\phi ([\vx_1,\cdots,\vx_{T}]))\,,
\end{align*}
where ${\rm CNN}_\phi$ represents a 2D CNN with parameters $\phi$ and ${\rm max\_pool}$ is a 1D max pooling operator. Combined, they return a fixed dimensional feature vector $\vh$ that replaces $[\mathbf{e}_1^T, \; \mathbf{e}_2^T]^T$ as input to the classifier.
For each of these two models, we learn its parameters jointly with the classification task using the cross-entropy loss that we described in Section~\ref{sec:classification}.

Overall, we test five different methods for the math operation classification and feedback classification tasks. 
The first three methods use different encoding methods in conjunction with a classifier: i) using the GRU encoder to encode math expressions as input to the classifier, which we dub \textbf{GRU+C}, ii) using the tree embedding-based encoder instead, which we dub \textbf{TE+C}, and iii) using the CNN encoder instead, which we dub \textbf{CNN+C}. 
These methods do not learn explicit representations of math operations. 
The next two methods use the TransE and TransR frameworks to learn these representations using tree embeddings: iv) using tree embedding-based encoder as input to the TransE framework in conjunction with a nearest neighbor classifier, which we dub \textbf{TE+TransE}, and v) using the TransR framework instead of the TransE framework to study math operations in multiple relation spaces, which we dub \textbf{TE+TransR}. 

\subsection{Experimental Setup}

We first test our math operation representation learning methods on the \texttt{OK} subset via 5-fold cross-validation, i.e., training on $80\%$ of steps in the subset to learn representations of math operations and testing them on the remaining $20\%$. 
We also test the generalizability of the learned representations to incorrect steps, i.e., replace the test set with the \texttt{ERROR} and \texttt{BUG} subsets, and check whether we can still recognize the math operation a student applied in an incorrect step. The results are detailed in Section~\ref{sec:1}. 

Since the distribution of math expressions in the \texttt{OK}, \texttt{ERROR} and \texttt{BUG} data subsets are mostly similar with minor differences, the previous experiment does not give us a good idea on the generalization ability of our math operation representation learning methods. 
Therefore, we further divide the \texttt{OK} subset into six smaller subsets, each corresponds to a different difficulty level (with different structure and complexity) according to questions within it, and test the generalizability of the learned math operation representations. 
The results are detailed in Section~\ref{sec:2}.
In practice, solution step data generated by real students is often limited.
Therefore, we conduct two more experiments to test whether synthetically generated steps can help us learn math operation representations that generalize to real data.
First, we repeat the experiments above using synthetically generated steps as the training set. 
This synthetic training set consists of $1,000$ steps for each math operation defined in Table~\ref{table:steps} (adding up to a total of $7,000$ across different difficulty levels). 
The results are detailed in Section~\ref{sec:3}.
Second, to study the impact of synthetically generated data when real data is limited, we pre-train the math operation representations with synthetic data, fine-tune on a small amount of real data from each difficulty level in the \texttt{OK} subset, and test on the rest. The results are detailed in Section~\ref{sec:4}. 

To test the ability of our learned math operation representations on recognizing student errors, we use them to classify feedback types provided by CogTutor in the \texttt{BUG} data subset.
Examples of such errors include when a student calculated the wrong simplification result, used the wrong sign in front of terms, and applied useless/unlogical steps to solve the problem, etc. 
The results are detailed in Section~\ref{sec:5}.

We use Adam optimizer~\cite{kingma2014adam} with learning rate $0.001$, batch size $64$ and run $10$ training epochs for each experiment. The math expression encoder outputs length-512 embedding vectors for each math expression, which we adapt to length-32 embedding vectors dimensions using a trainable fully-connected neural network. All of our experiments were conducted on a server with a single Nvidia RTX8000 GPU.

\subsection{Results and Discussion}

\subsubsection{Generalizing to incorrect steps}
\label{sec:1}
\begin{table}[t]
    \centering
    \resizebox{\columnwidth}{!}{
        \begin{tabular}{l|l|l|l}
            \toprule
            & \texttt{OK} & \texttt{ERROR} & \texttt{BUG} \\
            \midrule
            GRU+C & $99.18 \pm 0.23$ & $93.87 \pm 0.66$ & $95.89 \pm 0.63$ \\ \hline
            TE+C & $99.82 \pm 0.04$ & $93.30 \pm 0.65$ & $95.38 \pm 0.62$ \\ \hline
            CNN+C & $95.37 \pm 0.44$ & $86.82 \pm 1.38$ & $91.02 \pm 0.59$ \\ \hline
            TE+TransE & $96.27 \pm 0.17$ & $86.32 \pm 1.23$ & $84.21 \pm 2.13$ \\ \hline
            TE+TransR & $99.17 \pm 0.21$ & $91.28 \pm 1.12$ & $91.31 \pm 1.87$ \\
            \bottomrule
        \end{tabular}
    }
    \caption{Math operation classification accuracy for all methods training on the \texttt{OK} subset of the CogTutor dataset and testing on different data subsets. Accuracy is high across the board, while GRU-based encoding and tree embedding-based encoding in conjunction with a classifier result in the best performance.}
    \label{tbl:real}
\end{table}

Table~\ref{tbl:real} shows the averages and standard deviations of math operation classification accuracy for every method we experimented with using the \texttt{OK} subset as the training set. 
As expected, testing on the \texttt{ERROR} and \texttt{BUG} subsets result in slightly lower ($5$-$10\%$) math operation classification accuracy for all methods since the training set does not contain incorrect steps. 
However, even on steps that are incorrect, these methods can still effectively identify the math operation a student intended to apply (with up to $95\%$ accuracy), suggesting that they may be applicable to fully open-ended question solving solutions that are not highly structured, unlike those in Cognitive Tutor, to provide feedback to teachers on students' solution approaches. 

We observe that using GRUs and tree embeddings as representations for math expressions and applying a classification method on top of these representations result in similar performances; GRUs slightly outperform tree embeddings in cases where we use the \texttt{ERROR} and \texttt{BUG} subsets as the test set while tree embeddings slightly outperform GRUs in the case where we use a part of the \texttt{OK} subset as the test set. 
Using CNNs to encode math expressions as input to a classifier results in worse performance, suggesting that they do not capture the semantic and structural information in math expressions as well as GRUs and tree embeddings. 
As expected, using tree embeddings under the TransE and TransR frameworks leads to worse performance than the first two methods, with TransE achieving low performance (especially on the \texttt{BUG} subset) and TransR achieving comparable performance to the classification-based methods on the \texttt{OK} subset but lower performance on the \texttt{ERROR} and \texttt{BUG} subsets.
This result can be explained by the additional structural restriction that math operations are represented as linear and additive in some embedding space in the TransE framework, which makes it less robust against incorrect student solution steps. 
Using the TransR framework mitigates this problem due to its use of different relation spaces for each math operation. 

These methods perform similarly in the math operation classification task on real data largely due to the limited variation and complexity in the math expressions. 
The Cognitive Tutor system limits the degrees of freedom in a students' response by splitting an open-ended step into the separate actions of selecting a single math operation and entering the resulting math expression, which limits the variability in the data.
In the next experiment, we see that when we control against different levels of complexity in these math expressions and forcing these methods to generalize across complexities, their performance vary significantly.

\begin{figure}[t!]
    \centering
    \begin{subfigure}[b]{0.4\textwidth}
    \includegraphics[width=\linewidth]{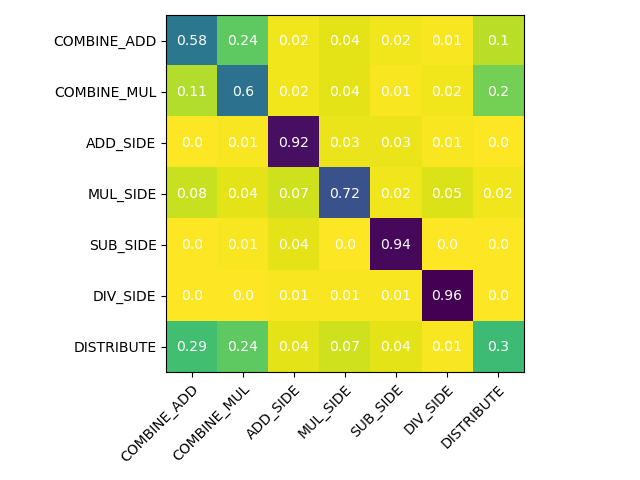}
    \vspace{-15pt}
    \caption{\centering Confusion matrix of math operations classification.}
    \end{subfigure}
    
    \vspace{10pt}
    \begin{subfigure}[b]{0.4\textwidth}
    \includegraphics[width=\linewidth]{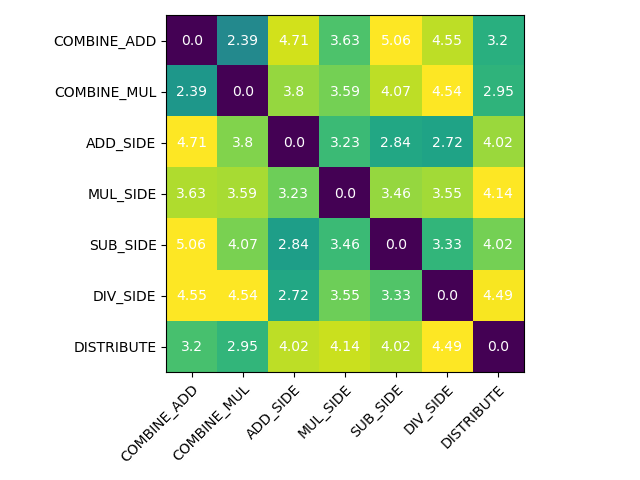}
    \vspace{-15pt}
    \caption{\centering Euclidean distance between learned math operation embedding vectors.}
    \end{subfigure}
    \caption{Details of TE+TransE for the math operation classification task on the \texttt{OK} subset. 
    These results match our intuition on how these math operations are related.
    }
    \label{fig:qualitative-realdata}
\end{figure}

\begin{figure}[t!]
 \vspace{-0.5cm}
    \centering
    \includegraphics[width=.45\textwidth]{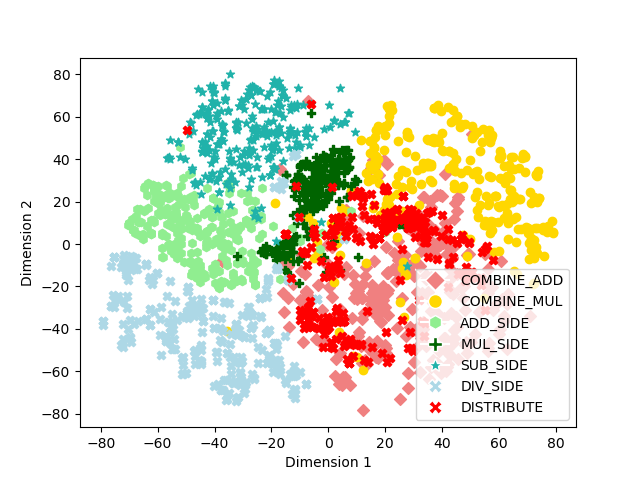}
    \vspace{-0.0cm}
    \caption{Visualization of learned math expression change for a randomly sampled subset of student solution steps in 2-D and corresponding operations (best viewed in color).}
    \label{fig:visual}
\end{figure}

Figure~\ref{fig:qualitative-realdata} visualizes the confusion matrix for math operation classification on the \texttt{OK} subset and the pairwise euclidean distances between math operation embeddings learned via the TransE framework using tree embeddings for math expressions. 
Rows correspond to the true math operations applied in steps and columns correspond to predicted ones. 
Percentages in the confusion matrix (Figure~\ref{fig:qualitative-realdata}a) are normalized w.r.t.\ the number of appearances of each math operation. 
We see that our math operation representation learning method captures some meaning of these operations (Figure~\ref{fig:qualitative-realdata}b); the learned math operation embeddings capture the structural changes in math expression in ways that match our intuition.
For instance, both COMBINE\_ADD and COMBINE\_MUL can be considered types of simplifications, so the Euclidean distance between the learned embeddings for these two operations is low.
This observation is not surprising due to the similar nature of these operations. Moreover, COMBINE\_ADD, COMBINE\_MUL, and DISTRIBUTE are often confused with one another. 
These results are also validated by a 2-D visualization (using t-SNE \cite{tsne} as a dimensionality reduction method) of the learned math operation embeddings in Figure~\ref{fig:visual}, where different math operations are mostly well separated except for COMBINE\_ADD, COMBINE\_MUL, and DISTRIBUTE. 
One possible explanation is that these operations are all applied to one side of the equation during a solution step, leaving one side of the equation unchanged, while the other operations, such as ADD\_SIDE, SUB\_SIDE, MUL\_SIDE, and DIV\_SIDE are all applied to both sides of the equation. 
Therefore, this result suggests that tree embeddings enable us to characterize a math operation by the \emph{structural} change in math expressions before and after a solution step where it is applied. 
Furthermore, the classification accuracy for the DISTRIBUTE operation is significantly lower than that for other operations. 
This result is likely due to the fact that the number of steps with this operation is significantly lower than that for other operations. 

\subsubsection{Generalizing to different difficulty levels}
\label{sec:2}

In this experiment, we test the ability of our learned math operation representations to generalize to math expressions with different levels of complexity in questions at different levels of difficulty. 
Although they are all about equation solving, questions at different difficulty levels in Cognitive Tutor involve math expressions that look very different. 
For example, in the easiest level (ES\_01), the equation that needs to be solved in a question looks like $x+5=9$, with only a single variable and without numbers with decimals. 
In contrast, in the hardest level (ES\_07), a question may contain coefficients with several decimal places and multiple variables, such as solve for $m$ in the equation $m(k-n)=gs$. 
We only compare the GRU-based encoder and the tree embedding-based encoder in conjunction with a classifier since they are the best performing methods in the previous experiment.
Table~\ref{table:level} lists the math operation classification accuracy for both methods after training on steps at different difficulty levels in the \texttt{OK} subset and testing on steps at other difficulty levels (including incorrect ones).
We see that TE+C overall outperforms GRU+C in almost every case. 
This results suggest that tree embeddings are effective at capturing the structural property of a math expression.
As a result, math operation representations based on tree embeddings excel at capturing the structural \emph{change} in math expressions before and after applying a math operation, leading to better generalizability than GRU-based encoding that do not explicitly account for this change. 

\begin{table}[t!]
    \centering
    \resizebox{\columnwidth}{!}{
        \begin{tabular}{c|l|l|l|l}
            \toprule
            \makecell{Train \\ on} & Method & \texttt{OK} & \texttt{ERROR} & \texttt{BUG} \\
            \midrule
            \multirow{2}{*}{ES\_01} &
            GRU+C & $58.82 \pm 1.12$ & $63.74 \pm 1.13$ & $66.02 \pm 1.12$ \\ 
            & TE+C & $76.51 \pm 0.62$ & $84.24 \pm 0.87$ & $67.49 \pm 1.10$ \\ \hline
            
            \multirow{2}{*}{ES\_02} &
            GRU+C & $71.05 \pm 1.12$ & $76.66 \pm 1.11$ & $69.01 \pm 1.14$ \\ 
            & TE+C & $87.89 \pm 0.34$ & $93.96 \pm 0.72$ & $80.44 \pm 0.78$ \\ \hline
            
            \multirow{2}{*}{ES\_03} &
            GRU+C & $82.39 \pm 3.93$ & $79.24 \pm 1.47$ & $80.01 \pm 1.67$ \\ 
            & TE+C & $90.79 \pm 1.12$ & $93.83 \pm 1.32$ & $84.70 \pm 1.54$ \\ \hline
            
            \multirow{2}{*}{ES\_04} &
            GRU+C & $76.72 \pm 0.14$ & $71.35 \pm 6.12$ & $83.32 \pm 2.24$ \\ 
            & TE+C & $94.65 \pm 0.12$ & $92.72 \pm 1.32$ & $90.99 \pm 1.72$ \\ \hline
            
            \multirow{2}{*}{ES\_05} &
            GRU+C & $81.74 \pm0.33 $ & $73.36 \pm 1.69$ & $78.36 \pm 1.07$ \\ 
            & TE+C & $87.66 \pm 0.25$ & $80.00 \pm 1.32$ & $77.81 \pm 0.99$ \\ \hline
            
            \multirow{2}{*}{ES\_07} &
            GRU+C & $76.25 \pm 3.21$ & $73.15 \pm 3.42$ & $67.35 \pm 3.62$ \\ 
            & TE+C & $79.44 \pm 0.62$ & $79.29 \pm 0.72$ & $72.53 \pm 2.26$ \\
            \bottomrule
        \end{tabular}
    }
    \caption{Math operation classification accuracy after training on steps with different difficulty levels and testing on the \texttt{OK} \texttt{ERROR}, and \texttt{BUG} subsets. Tree embedding-based encoding outperforms GRU-based encoding.}
    \label{table:level}
\end{table}

\subsubsection{Generalizing to different data distributions}
\label{sec:3}
\sloppy
In this experiment, we test the ability of our methods to generalize from synthetically generated data to real student data. 
We train different math operation classification methods on the $2,000$ synthetically generated steps and test them on steps generated by real students in the CogTutor dataset. 
Table~\ref{table:synthetic} shows the mean and standard deviation for each method on each real data subset. 
We see that TE+C significantly outperforms GRU+C and CNN+C on all data subsets, which is in stark contrast to the previous experiment where the difference in performance across all methods is much smaller. 
This observation suggests that tree embeddings are more effective at capturing the semantic/structural effect of math operations on math expressions, thus generalizing better to different data distributions. 
Indeed, although the synthetically generated steps and the real steps have the same set of math operations, the distributions of numbers ($1,0.5,-7$, etc.) and variables ($x,u,t$, etc.), resulting in a mismatch between the data distributions. 
Tree embedding-based methods benefit from the tree-based representations of math expressions that can effectively capture structural information, making it easy for the learned embeddings of math expressions to generalize to unseen data. 

\begin{figure}[t!]
    \centering
    \includegraphics[width=.4\textwidth]{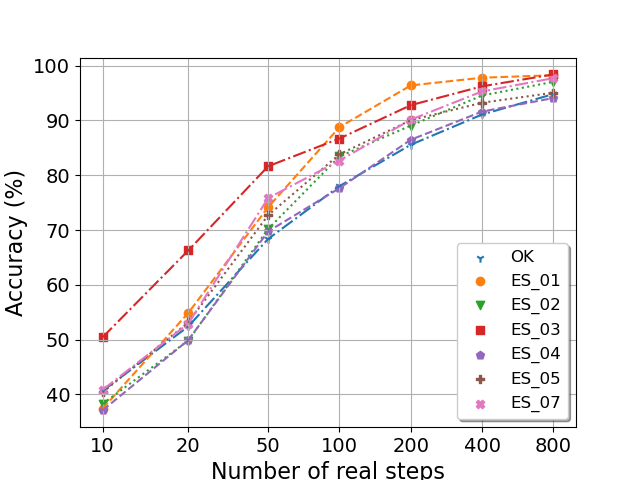}
    \caption{Math operation classification accuracy for the TE+C method when real data is limited. Using synthetically generated steps as a starting point, we already start with acceptable classification accuracy even with few real steps generated by students. The performance steadily improves after more real data becomes available.}
    \label{fig:small_data}
\end{figure}

\begin{table}[t!]
    \centering
    \resizebox{\columnwidth}{!}{
        \begin{tabular}{l|l|l|l}
            \toprule
            & \texttt{OK} & \texttt{ERROR} & \texttt{BUG} \\
            \midrule
            GRU+C & $62.89 \pm 3.93$ & $64.06 \pm 4.70$ & $62.94 \pm 2.24$ \\ \hline
            TE+C & $83.79 \pm 0.14$ & $75.49 \pm 0.90$ & $75.16 \pm 0.55$ \\ \hline
            CNN-C & $51.12 \pm1.64 $ & $45.52 \pm 0.98$ & $59.82 \pm 1.68$ \\ \hline
            TE + TransE & $80.17 \pm 2.32$ & $71.86 \pm 3.24$ & $72.32 \pm 2.72$ \\ \hline
            TE + TransR & $82.22 \pm 2.88$ & $73.83 \pm 3.46$ & $74.85 \pm 3.23$ \\
            \bottomrule
        \end{tabular}
    }
    \caption{Math operation classification accuracy for all methods training on $7,000$ synthetically generated steps and testing on different subsets of the CogTutor dataset. Tree embedding-based methods significantly outperform other methods, showing better ability to generalize to different data distributions.}
    \label{table:synthetic}
\end{table}

\subsubsection{Generalizing from synthetic data}
\label{sec:4}
\begin{figure}[t]
    \centering
    \includegraphics[width=.4\textwidth]{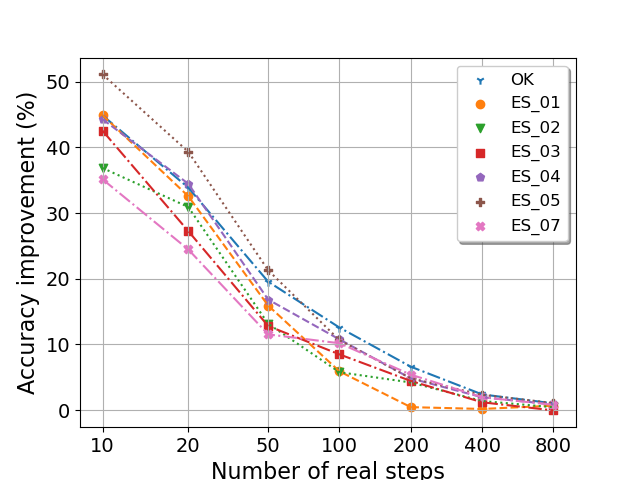}
    \caption{Math classification accuracy (difference in percentage) for TE+C,  pre-training on synthetic data before fine-tuning on real data versus training only on real data. When real data is limited, pre-training on synthetic data results in significantly better performance.}
    \label{fig:fine_tuned}
\end{figure}

Ideally, if there is a large amount of training data, i.e., steps generated by real students containing different types of math expressions and detailed labels on these steps such as the math operation(s) applied, the error(s) if a step is incorrect, and corresponding feedback, we can simply use that data to learn our math operation representations. 
However, in practice, the amount of real data is often limited.
Figure~\ref{fig:small_data} plots the performance of TE+C on all subsets of the CogTutor dataset, training on a portion of steps in the subset for training and testing on the rest.
We see that the performance on math operation classification suffers considerably when we only have limited training data. 
Therefore, synthetically generated data can play a vital role in improving their performance under this circumstance; the strategy of fine-tuning models trained on synthetically generated data using a small amount of real data can be effective. 
Specifically, we start with a pre-trained math operation classification model on the $7000$ synthetically generated steps and fine tune it on a small number of real steps by doing gradient descent on these steps for $10$ epochs. 
Figure~\ref{fig:fine_tuned} plots the improvement in math operation classification accuracy for the fine-tuned model over the model that trains on only real data of various amounts on all data subsets. 
We see that the pre-trained models always performs better, with significant improvement when the real data is extremely limited. 
This result suggests that i) effectively leveraging synthetically generated data can mitigate the problem of limited real data and ii) our math operation representation learning methods are capable of generalizing across different data distributions (synthetic $\rightarrow$ real).

\subsubsection{Feedback type classification}
\label{sec:5}

In this experiment, we evaluate our math operation representation learning methods on the feedback type classification task. 
These feedback items were automatically deployed by Cognitive Tutor for incorrect steps in the \texttt{BUG} subset. 
We pre-processed these steps and grouped the detailed feedback items according to the students' errors that each feedback item addresses and narrowed it down to a total of $24$ types that occur multiple times. 
We perform 5-fold cross validation on this subset. 
Table~\ref{table:feedback} shows the averages and standard deviations of feedback classification accuracy for all methods on this task across the five folds. 
We see that due to the limited size of the \texttt{BUG} subset (only $5,744$ steps) and the high number of classes ($24$), all method perform worse than they do on the math operation classification task. 
Specifically, we see that the tree embedding-based encoder in conjunction with a classifier performs best while GRU-based encoding also performs well. 
This result shows that although tree embeddings are superior at capturing the meaning of math expressions, their advantage over simple encoding methods such as GRU-based encoding decreases due to increased noise in the data; some math expressions submitted by students in incorrect steps are ill-posed and do not make sense. 
Using the TransE and TransR frameworks result in slightly worse performance than classifiers since these methods explicitly learn a representation for each math operation, which limits their performance on this task due to the shortage of training data.
However, since they capture the structural difference in math expressions before and after the step, they can cancel out some of the noise in erroneous steps, resulting in acceptable performance. 

\begin{table}[t!]
    \centering
        \begin{tabular}{l|l}
            \toprule
            \textbf{Method} & \textbf{Accuracy} \\
            \midrule
            GRU+C & $75.35 \pm 1.41$ \\ \hline
            TE + C & $78.71 \pm 1.74$\\ \hline
            CNN+C & $67.23 \pm1.54 $  \\ \hline
            TE + TransE & $69.15 \pm 1.13$ \\ \hline
            TE + TransR & $73.21 \pm 1.63$  \\ 
            \bottomrule
        \end{tabular}
    \caption{\sloppy Feedback type classification accuracy for all methods on the \texttt{BUG} subset. Tree embedding-based encoding outperforms other encoding methods while TransE and TransR frameworks do not reach similar performance levels due to shortage of training data.}
    \label{table:feedback}
\end{table}

\subsection{Discussions}
Overall, we find that the GRU-based and tree embedding-based math expression encoders in conjunction with a classifier perform almost equally well in most situations, while the CNN-based encoder performs worse. 
The tree embedding-based encoder has stronger generalizability across different data distributions. 
We believe that as the math expressions and operations get more complicated, methods that leverage the tree structure of math expressions would be more advantageous. 
We also observe that TransR outperforms TransE most of the time, although in some experiments using TransE and TransR to explicitly learn math operation embeddings lead to slightly worse performance than classifiers using implicit representations of math expressions. 
However, TransE and TransR are much more powerful and enable us to study more tasks such as clustering solution steps and identifying typical student errors and learning solution \emph{strategies}; See Section~\ref{sec:conc} for a detailed discussion. 

\section{Conclusions and Future Work}
\label{sec:conc}

In this paper, we developed a series of methods to learn representations of math operations by observing how math expressions change as a result of these operations in step-by-step solutions to open-ended math questions. 
Our methods leverage math expression encoding methods that map tree-structured math expressions into a math embedding vector space. 
We demonstrated the effectiveness of our methods on a dataset containing detailed student solution steps to equation solving questions in the Cognitive Tutor system on two tasks: i) classifying the math operation applied in each step and ii) classifying the feedback the system deploys for each incorrect step. 
Results show that our learned math operation representations are meaningful and can often effectively generalize across different data distributions such as questions with different difficulty levels.

However, the success of our methods heavily depends on the availability of diverse large-scale training data. 
The Cognitive Tutor dataset that we used in this work represents a heavily restricted solution process since the list of math operations a student can apply in a step is pre-defined. 
Therefore, additional work has to be done to extend our method to truly open-ended step-by-step solution processes that are less structured. 
Moreover, our methods are restricted to a single solution step only and do not consider the relationship across multiple steps, which is related to another important aspect of solving open-ended math questions: the overall solution strategy, i.e., which math operation to apply next. 
Furthermore, in both classification tasks, using tree embeddings to encode math expressions in conjunction with a classifier outperforms explicitly learning vectorized representations of math operations in the TransE and TransR frameworks. 
However, these explicit representations may enable us to perform other tasks such as 
Nevertheless, our work provides a series of tools to analyze the math expressions students write down in their solutions by bridging the gap between symbolic math representations with continuous representations in vector spaces, enabling the use of state-of-the-art neural network-based methods. 
We believe that this work can potentially open up a new line of research that studies how to automatically analyze student solutions for grading and feedback purposes. 

There are many avenues of future work. 
First, since most real-world open-ended solutions contain a mixture of math expressions and text, there is a need to learn a joint representation of math expressions and text in a shared embedding space. 
Second, this joint representation will enable us to train automated feedback generation methods in an end-to-end manner, using sequence-to-sequence learning methods \cite{seq2seq}. 
Third, using learned math expression representations as the states and learned math operation representations from the TransE and TransR frameworks as the state transition model, we can apply reinforcement learning and inverse reinforcement learning methods to learn solution strategies, i.e., which math operation to apply in the next step. 
We can also study solution strategies employed by real students \cite{rafferty} and diagnose their errors and design corresponding feedback mechanisms to improve their learning outcomes. 
These future work directions will enable us to tap into the full potential of explicit math operation representations, which is not fully demonstrated in this paper: on the CogTutor dataset, the only relevant real-world dataset we found, we could only evaluate these explicit representations on the math operation and feedback prediction tasks, where they may not outperform tree embedding-based classification-based methods. 
\newpage
\balance
\bibliographystyle{abbrv}
\bibliography{main}
\end{document}